\documentclass{supcon}

\usepackage[T1]{fontenc}
\usepackage[utf8]{inputenc}
\usepackage{graphicx}
\usepackage{amsmath}
\usepackage{microtype}
\usepackage{hyperref}
\usepackage{booktabs}
\usepackage{multirow}
\usepackage{xcolor}
\usepackage{xspace}
\usepackage{subfigure}

\makeatletter
\providecommand{\@LN@col}[1]{}
\providecommand{\@LN}[2]{}
\providecommand{\@LN@postlabel}[1]{}
\makeatother

\definecolor{supconGray}{RGB}{248,250,252}

\newcommand{\ours}{HEAR\xspace}
\newcommand{\dofba}{OF-BRCA\xspace}
\newcommand{\dofgeneral}{OF-General\xspace}
\newcommand{\supcon}{SUPCON\xspace}
\newcommand{\para}[1]{\smallskip\noindent\textbf{#1}}
\newcommand{\todo}[1]{\xspace}
\newcommand{\todowl}[1]{\xspace}
\newcommand{\fix}[1]{\xspace}

\setsupconlogo{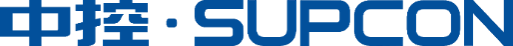}
\makeatletter
\renewcommand{\supcon@logobox}{%
	\IfFileExists{\supconlogo}{%
		\includegraphics[height=0.28cm]{\supconlogo}%
	}{%
		{\color{supconBlue}\textbf{SUPCON}}%
	}%
}
\makeatother

\makeatletter
\g@addto@macro\maketitle{\par\@thanks}
\makeatother
\setsupconcorrespondence{\mbox{Ling Wang(wangling3@supcon.com)}}

\title{\raggedright Hypergraph Enterprise Agentic Reasoner over Heterogeneous Business Systems}
\author{Ling Wang$^{*}$, Xin Liu$^{*}$, Songnan Liu, Jianan Wang, Cheng Cheng, Yihan Zhu, Enyu Li, Yu Xiao, Jiangyong Xie, Duogong Yan, Jiangyi Chen\\
SUPCON, Hangzhou, China\\
\{wangling3,liuxin4,liusongnan,wangjianan,chengcheng1,\\
zhuyihan,lienyu,xiaoyu2,xiejiangyong,yanduogong,chenjiangyi\}@supcon.com}
\date{}

\supconabstract{%
Applying Large Language Models (LLMs) to heterogeneous enterprise systems is hindered by hallucinations and failures in multi-hop, n-ary reasoning. Existing paradigms (e.g., GraphRAG, NL2SQL) lack the semantic grounding and auditable execution required for these complex environments. We introduce \ours, an enterprise agentic reasoner built on a Stratified Hypergraph Ontology. Its base Graph Layer virtualizes provenance-aware data interfaces, while the Hyperedge Layer encodes n-ary business rules and procedural protocols. Operating an evidence-driven reasoning loop, \ours dynamically orchestrates ontology tools for structured multi-hop analysis without requiring LLM retraining. Evaluations on supply-chain tasks, including order fulfillment blockage root cause analysis (RCA), show \ours achieves up to 94.7\% accuracy. Crucially, \ours demonstrates adaptive efficiency: utilizing procedural hyperedges to minimize token costs, while leveraging topological exploration for rigorous correctness on complex queries. By matching proprietary model performance with open-weight backbones and automating manual diagnostics, \ours establishes a scalable, auditable foundation for enterprise intelligence.
}

\begin{document}
\maketitle
\begingroup
\renewcommand{\thefootnote}{*}
\footnotetext{These authors contributed equally to this work.}
\endgroup
\setcounter{footnote}{0}

\section{Introduction}

\begin{figure}[t]
    \centering
    \includegraphics[ width=\textwidth, trim=0cm 10cm 1cm 0cm, 
    clip]{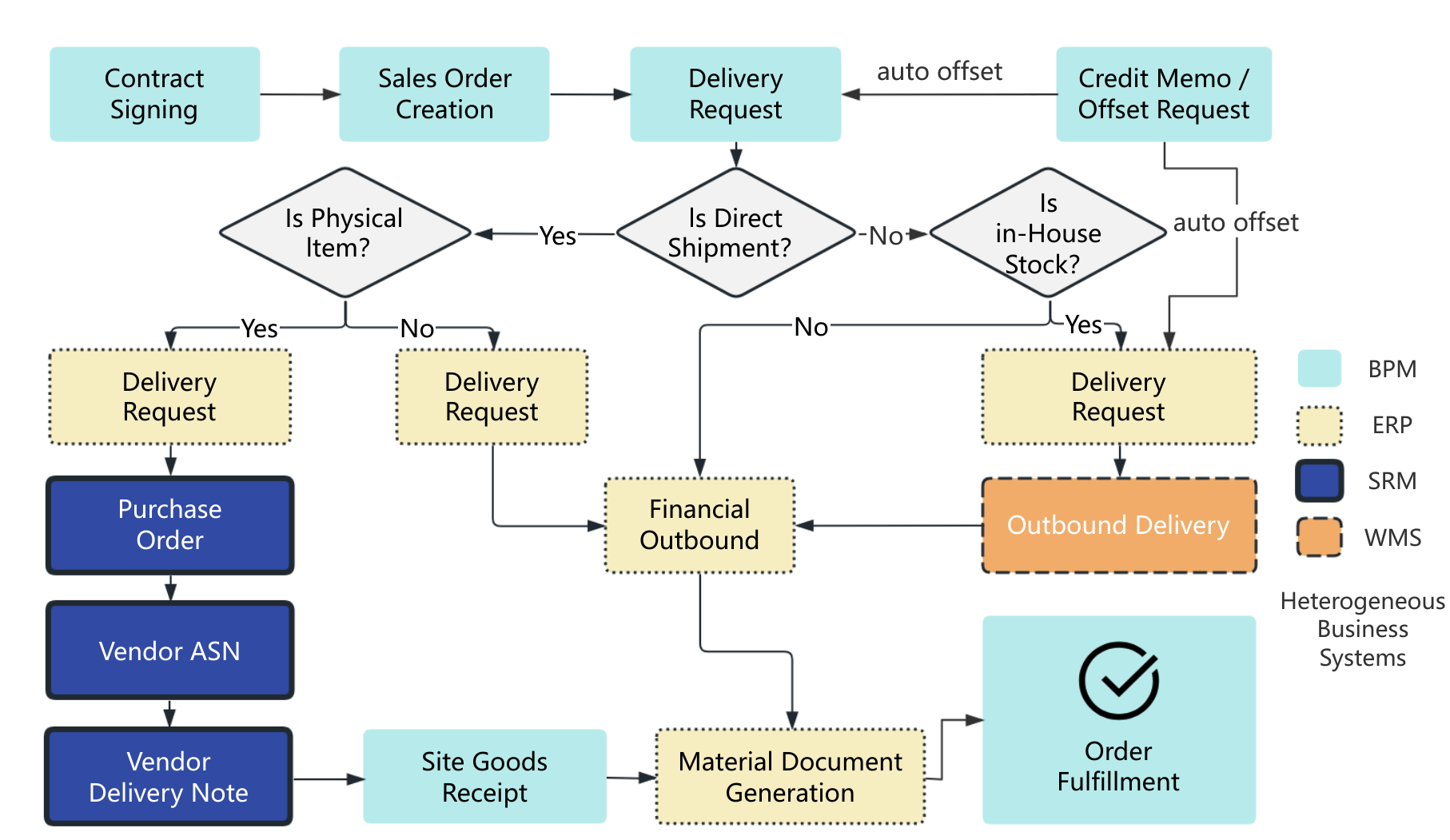}
    \caption{Order fulfillment workflow across four heterogeneous business systems: BPM, ERP, SRM, and WMS in \supcon.}
    \label{fig:of-workflow}
\end{figure}

Modern businesses rely on heterogeneous systems—including Enterprise Resource Planning (ERP) \cite{kurbel2013erp}, Supplier Relationship Management (SRM) \cite{schuh2014srm}, Warehouse Management Systems (WMS) \cite{bartholdi2016warehouse}, and Business Process Management (BPM) \cite{books/sp/DumasRMR18}—to manage end-to-end workflows like order fulfillment (Figure~\ref{fig:of-workflow}). While providing robust transactional capabilities, they often lack semantic interoperability due to severe data fragmentation. Real-world tabular data is notoriously noisy and misaligned, exhibiting schema ambiguity, inconsistent identifiers and formats, and opaque statistical metrics. For instance, custom development on a German SAP system within a Chinese enterprise often yields schema ambiguities, confusing original abbreviations with localized Pinyin acronyms.

Beyond data-level misalignment, these systems are governed by complex business constraints that span multiple platforms. These constraints dictate conditional workflow branching (e.g., routing in-house stock to delivery requests on ERP and then outbound delivery on WMS, versus directing direct-shipment physical items to SRM for procurement as shown in Figure \ref{fig:of-workflow}), status transitions, and cross-system synchronization (e.g., aligning BPM delivery requests with ERP material document generation). Together, these factors make routine analytical tasks such as fulfillment diagnosis and delay tracing a coupled challenge of cross-system evidence acquisition and business-logic reasoning, where relevant facts are distributed, only partially aligned, and constrained by domain-specific rules \cite{antelmi2023survey,luo2025hypergraphrag}.

Although Large Language Models (LLMs) excel at natural language reasoning, they are severely hampered by this lack of semantic interoperability when navigating heterogeneous multi-hop tasks and complex business constraints. 
If forced to rely merely on misaligned pre-trained knowledge, they frequently suffer from severe factual hallucinations and ``reasoning stalls'' \cite{benchmarking2025deepsearch,xu-etal-2024-knowledge-conflicts}. Standard Retrieval-Augmented Generation (RAG) paradigms seem to be a promising solution, but they fail to capture the complex $n$-ary relational facts required for such decision support \cite{lewis2020rag,tang2024raglimitations}. Even GraphRAG \cite{han2025graphrag,graphrag_survey} is limited by binary structures that cannot adequately model $n$-ary constraints. Hyper-topology RAG \cite{luo2025hypergraphrag,xie2025hypkg,feng2025hyperrag} triggers index explosion, while NL2SQL \cite{li2023nlidb,lei2024spider2} and general agents \cite{biswal2026agentsm,zhai2025excot} lack the simultaneous semantic grounding, topology awareness, and auditable execution required for production-grade enterprise intelligence.

To bridge this gap, we propose \textbf{\ours} (\textbf{H}ypergraph \textbf{E}nterprise \textbf{A}gentic \textbf{R}easoner), an enterprise agentic reasoning system grounded in a \textbf{Stratified Hypergraph Ontology}. By representing complex, $n$-ary relationships as hyperedges, \ours provides a unified framework for multi-hop reasoning that seamlessly spans heterogeneous systems. In this architecture, 
the Graph Layer standardizes binary connections across heterogeneous data sources, providing a virtualized, provenance-aware factual substrate.
The Hyperedge Layer functions as an $n$-ary ontology whose hyperedges bind subsets of cross-domain graph nodes. Instead of requiring rigid symbolic representations, \ours uses human- and machine-interpretable natural-language soft axioms in \emph{declarative hyperedges} which ground admissible business states, cross-system relationships, and semantic interpretation in established graph nodes, while \emph{procedural hyperedges} build on these anchors to construct detailed execution protocols and reusable reasoning workflows for routine tasks. An orchestration LLM acts as the reasoning engine, dynamically activating these hyperedges and calling tools to navigate the underlying data topology while adhering to business constraints and validated protocols. Importantly, the system evolves through the continuous, human-in-the-loop construction of these hyperedges, securely adapting to new business logic without retraining the underlying LLM.

Empirically, on real-world supply-chain tasks, this hypergraph-grounded design yields up to 94.7\% accuracy on complex blockage analysis and 88.7\% on general questions. Crucially, \ours exhibits highly adaptive execution efficiency. On bounded diagnostic tasks, predefined procedural hyperedges actively constrain exploration, drastically reducing token consumption compared to baselines. Conversely, when forced into autonomous multi-hop exploration for diverse queries, \ours deliberately scales its computational investment to secure indispensable correctness, significantly widening its strict accuracy gap over baseline methods that severely degrade. Additionally, deploying \ours with cost-efficient open-weight models achieves proprietary-level performance and translates minimal ontology maintenance into a scalable paradigm shift that automates recurring manual diagnostics.

The main contributions of this work are summarized as follows:
\begin{itemize}
    \item We propose a \textbf{Stratified Hypergraph Ontology} for enterprise intelligence. The Graph Layer registers provenance-aware data interfaces over heterogeneous systems, while the Hyperedge Layer serves as an $n$-ary ontology whose declarative hyperedges bind localized soft axioms to graph nodes and whose procedural hyperedges construct reusable execution protocols over those grounded anchors. 

    \item We design \textbf{\ours}, an enterprise agentic reasoner with an evidence-driven agentic reasoning loop on top of the stratified hypergraph ontology. The LLM acts as the reasoning engine that performs semantic planning and invokes ontology tools for verifiable evidence acquisition and ontology building, supporting both procedural hyperedge-guided reuse and topology-guided generalization for long-tail tasks.

    \item We provide a \textbf{real-world empirical evaluation} on enterprise supply-chain tasks. Results demonstrate that explicit hypergraph semantics drive exceptional robustness and strict accuracy as reasoning spans increase. Notably, \ours exhibits adaptive execution efficiency: it minimizes token costs via predefined procedural hyperedges on standard tasks, while deliberately trading raw compute for indispensable correctness during multi-hop exploration. Furthermore, it achieves proprietary-level accuracy using cost-efficient open-weight models, automating cross-system manual diagnostics with minimal maintenance overhead.
\end{itemize}

\section{Related Work}

The transition toward enterprise-grade intelligent systems intersects with several research areas aimed at enhancing LLMs for enterprise applications.

\para{Hallucination Mitigation and the Need for Grounding.} LLM hallucinations hinder enterprise adoption \cite{cossio2025taxonomy,ji2023surveyhallucination,alansari2025hallucination}, particularly within complex, out-of-distribution business logic \cite{tan2023chatgpt,ma2025llmkg4qa}. Treating LLMs as monolithic reasoners causes fragile logical chains, ``reasoning stalls,'' and factual fabrication \cite{zaharia2024compound,yao2023react,xu-etal-2024-earth}. While semantic entropy detects uncertainties \cite{farquhar2024semantic}, it fails to resolve underlying knowledge conflicts. To effectively mitigate this, production-grade systems demand both \emph{informational grounding} for factual accuracy and \emph{structural grounding}: explicit topological constraints, verifiable tool-use, and auditable protocols \cite{fitzpatrick2026enterprise,rath2025governance,dutta2026pager}. Consequently, recent research explores retrieval- and graph-augmented architectures to achieve this, despite distinct bottlenecks.

\para{RAG and Graph-Augmented Methods.} While RAG grounds LLMs \cite{lewis2020rag}, standard single-hop retrieval struggles with multi-hop relationships across heterogeneous enterprise systems \cite{benchmarking2025deepsearch,koubarakis2025knowwheregraph}. GraphRAG and KBQA improve multi-hop reasoning \cite{graphrag_survey,mavromatis-karypis-2025-gnn,xu2024generate,liu2025ort} but incur high maintenance costs for indexing as data evolves \cite{graphrag_survey,han2025graphrag}. Conversely, Virtual Knowledge Graphs enable query without physical materialization \cite{Xiao2019Virtual,Calvanese2017Ontop,vidal2025neurosymbolic}. Other methods aggregate outputs via auxiliary graphs \cite{he2025llmforest}. Crucially, most graph systems assume binary triples, losing vital information when modeling $n$-ary enterprise events \cite{das2022cbr,zhang2022subgraph,antelmi2023survey}.

\para{Hyper-Relational Topology in RAG.} Binary graph structures often cause ``contextual fragmentation'' in multi-entity relationships \cite{han2025graphrag,antelmi2023survey}. Works like HyperGraphRAG and HypKG mitigate this by representing $n$-ary facts via hyperedges, preserving information density and beyond-pairwise correlations \cite{luo2025hypergraphrag,xie2025hypkg,feng2025hyperrag}. Recent advances further optimize hypergraph navigation: PRoH \cite{zai2026proh} dynamically plans multi-hop reasoning paths over knowledge hypergraphs, while Cog-RAG \cite{li2026cograg} utilizes dual-hypergraphs to achieve both global thematic and local entity alignment. However, applying these instance-level methods directly to enterprise tabular environments requires instantiating millions of tabular rows into a hypergraph, which could trigger index explosion, rendering them unsuitable for industrial environments.

\para{LLM-Driven Ontology and Constraint Modeling.} Ontologies are often constructed using formal semantic standards (e.g., RDF~\cite{cyganiak2014rdf} and OWL~\cite{hitzler2012owl}) alongside rigorous cross-system mappings. To streamline this modeling process, recent works explore LLM-driven generation: NeOn-GPT~\cite{fathallah2024neongpt} and OntoEKG~\cite{ontoekg2026} automate ontology creation; AutoQG~\cite{shen2026autoqg} and AttributeForge~\cite{huang2025attributeforge} scale schema generation via multi-agent orchestration; LLM-CG~\cite{liu2025llmcg} extracts relational constraints; and Matchmaker~\cite{seedat2025matchmaker} and ConStruM~\cite{chen2026construm} enable zero-shot schema matching. Inspired by recent advances~\cite{sharma2025ograg,zhang2025trustuqa}, \ours provides an alternative to these formal structural methods, utilizing natural language to explicitly represent $n$-ary axioms among nodes. Because this format is natively interpretable by both humans and LLMs, it establishes lightweight solutions that enable the agile, on-demand construction of semantic hypergraphs.

\para{Memory Systems and Agent Frameworks.} Memory systems extend context via structured representations \cite{packer2023memgpt,zhong2023memorybank,wang2025mirix,chhikara2025mem0,yang2026graphmemory}, but this passive factual recall often causes memory fragmentation by failing to enforce $n$-ary business constraints. Concurrently, unconstrained long interaction chains in agentic frameworks inevitably introduce semantic drift \cite{rath2026agentdrift,wegener2026sortai,dutta2026pager,maharaj2025evaluation}. \ours overcomes both bottlenecks via its Stratified Hypergraph Ontology. Instead of retrieving fragmented past interactions, \ours encapsulates semantics and task-specific execution protocols directly into hyperedges, providing topology-constrained pathways that actively guide tool invocation and prevent logical drift.

\para{Table-Aware Retrieval and Enterprise Querying.} NL2SQL techniques effectively bridge users and databases \cite{li2023nlidb} but struggle with the $n$-ary constraints and massive schemas typical of enterprise reasoning \cite{yu2018spider,lei2024spider2}. To handle heterogeneous tabular data, parallel table-centric RAG methods \cite{zou2025trag,yu2025tablerag} and schema-graph frameworks like CSR-RAG \cite{singh2026csrrag} attempt to map schemas for retrieval; however, they often falter on cross-system queries lacking explicit column-name linkage, which the hyperedges in \ours can supplement. Other approaches aim to improve robustness via interactive feedback \cite{menon2025fisql}, trace-reuse memory \cite{biswal2026agentsm,zhai2025excot}, or mapping schemas to retrievable text nodes (e.g., LlamaIndex \cite{llamaindex_sql_docs}). More critically, these isolated optimizations lack an iterative ReAct paradigm \cite{yao2023react} to enhance LLMs, so they remain unable to dynamically refine execution plans based on intermediate execution results—a structural gap that \ours's agentic reasoning loop explicitly addresses.

\para{Case Study: Supply-Chain Root-Cause Analysis.} Supply-chain root-cause analysis (RCA) inherently requires tracing multi-hop dependencies across highly heterogeneous enterprise systems (e.g., ERP, SRM, WMS, BPM) \cite{kurbel2013erp,schuh2014srm,bartholdi2016warehouse,kletti2007manufacturing}. To diagnose such complex anomalies, recent works integrate LLMs with knowledge graphs \cite{dixit2025pharmaRCA}, utilize ontology-guided extraction for risk monitoring \cite{wang2026supplyrisk,kosasih2025supplychain}, and explore hypergraph structures to model failure propagation \cite{xie2025root}. Addressing this systemic fragmentation, \ours specifically tackles order fulfillment blockage RCA by localizing stalled contracts through the composition of cross-system evidence. This concrete industrial application directly motivates our Stratified Hypergraph Ontology, providing the necessary structural foundation for topology-constrained retrieval and logical analysis.

\section{Problem Formulation}
\label{sec:problem_formulation}

To achieve intelligent operational decision analysis across heterogeneous enterprise systems, extremely stringent requirements \cite{benchmarking2025deepsearch} must be satisfied. To prevent the LLM from hallucinating, losing track of multi-hop dependencies, or stalling, the system must address three core challenges:

\begin{enumerate}

    \item \textbf{Data Fragmentation and Implicit Anomalies:} Enterprise knowledge is heavily fragmented across heterogeneous systems (e.g., ERP, WMS, SRM, BPM), where real-world tabular data is notoriously noisy and misaligned. To orchestrate reasoning, the engine must overcome severe \emph{semantic interoperability barriers} \cite{dingtalk2025deepresearch,Xiao2019Virtual}. It must dynamically resolve a myriad of implicit data and process anomalies---such as semantic ambiguity in schemas, field-level equivocation, opaque statistical metrics, and cross-system identifier inconsistencies.

    \item \textbf{Complex $n$-ary Business Constraints:} Workflows adhere to conditional logic involving multiple entities simultaneously. Capturing these multifaceted constraints requires representations beyond binary schema mappings to provide clear semantic and executive guidelines.
    
    \item \textbf{Verifiable Multi-Hop Reasoning:} Navigating topologies requires iterative reasoning to correct early hypothesis drift \cite{iterativerag2025}. To ensure audit compliance, every intermediate step and the final conclusion must be grounded in traceable data provenance.
\end{enumerate}

\paragraph{Formal Definition.} 

Given a natural language query $q$ representing the task, a heterogeneous data environment $\mathcal{D} = \{D_1, \dots, D_m\}$, and a set of soft axioms $\mathcal{R}$ encapsulating complex $n$-ary business constraints, the task is to orchestrate an LLM to dynamically generate a sequence of executable reasoning steps $\pi = (a_1, \dots, a_n)$. Each step $a_i$ may retrieve verifiable evidence from $\mathcal{D}$, apply relevant axioms in $\mathcal{R}$, or think about subsequent actions based on the current context. The ultimate objective is to synthesize a definitive response $r$, grounded in a completely auditable trace of data provenance and logical reasoning.

\section{Methodology}
\label{sec:methodology}

\begin{figure*}[t]
  \centering
  \includegraphics[ width=\textwidth, trim=3.3cm 0cm 3cm 2cm, 
        clip]{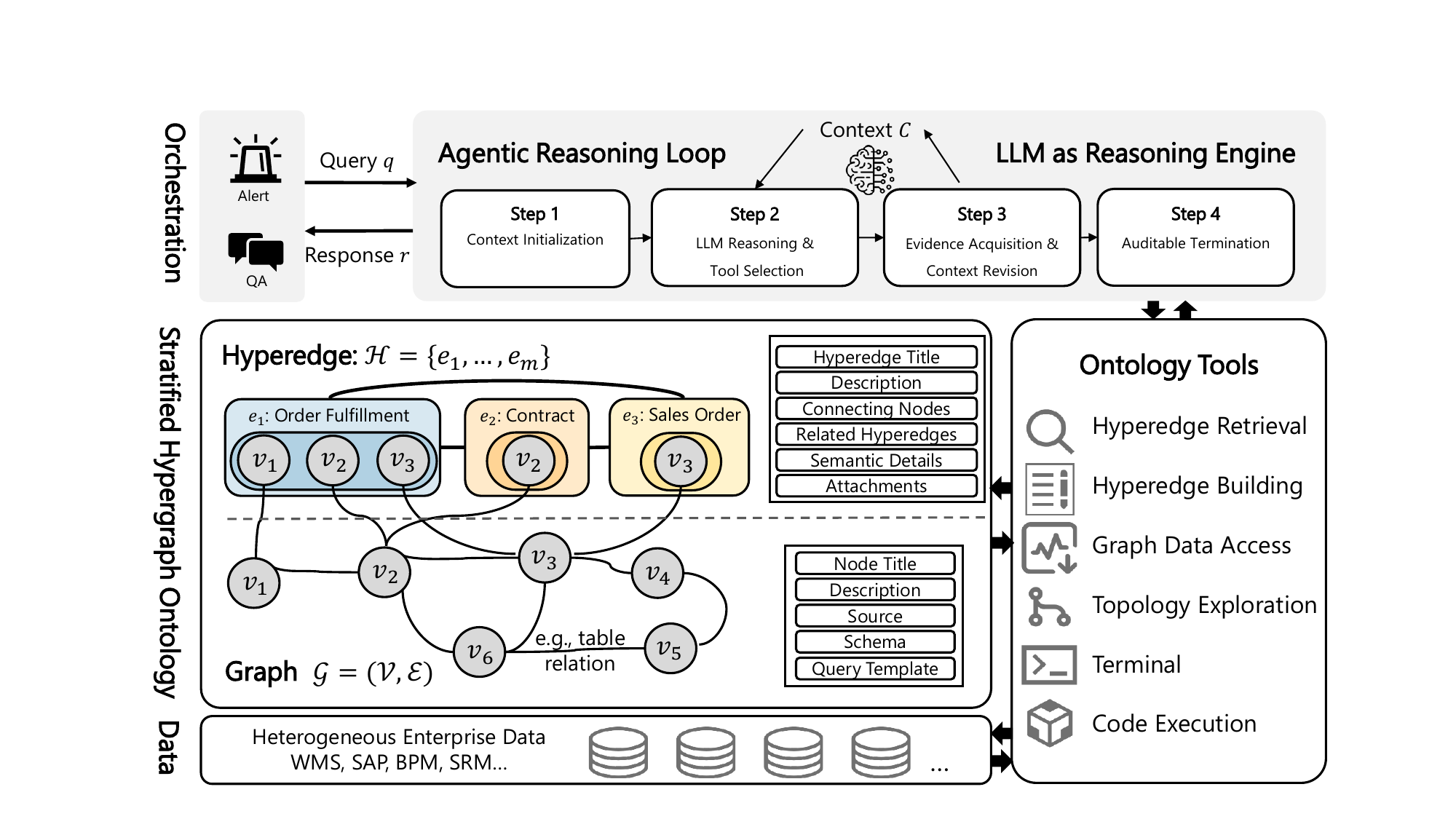}
  \caption{\textbf{Overview of \ours}. The system combines an Agentic Reasoning Loop with a Stratified Hypergraph Ontology: the Graph Layer registers virtualized, provenance-aware data interfaces and binary relations over heterogeneous business systems, while the Hyperedge Layer organizes $n$-ary semantic units on top of the base graph. An LLM acts as the reasoning engine and interacts with ontology tools over heterogeneous enterprise data.}
  \label{fig:framework}
\end{figure*} 

To address the multifaceted challenges of enterprise intelligence formulated in Section~\ref{sec:problem_formulation}, we propose \textbf{\ours}, an agentic reasoner built on a \textbf{Stratified Hypergraph Ontology}. As in Figure~\ref{fig:framework}, the \ours architecture operates as an end-to-end stack across three integrated layers. At the top, the \emph{Orchestration} layer hosts an LLM acting as the reasoning engine within an \emph{Agentic Reasoning Loop}, which iteratively drives context initialization, semantic planning, evidence acquisition, and auditable termination for a given query $q$. 
This loop is structurally grounded by the middle \emph{Stratified Hypergraph Ontology}, which separates a base \emph{Graph} $\mathcal{G}=(\mathcal{V},\mathcal{E})$ of virtualized, provenance-aware data interfaces as nodes from an upper \emph{Hyperedge} layer $\mathcal{H}$ that binds nodes via declarative $n$-ary soft axioms and procedural protocols.
The reasoning engine interacts with this substrate via a dedicated \emph{Ontology Tools} suite (enabling hyperedge retrieval, topology exploration, graph data access, and auxiliary executions), ensuring that reasoning plans are executed through verifiable operations. Finally, the foundational \emph{Data} layer encompasses heterogeneous enterprise systems (e.g., WMS, SAP, BPM, SRM) in-place. The following subsections formalize this ontology and the evidence-driven reasoning loop that operates upon it.

\subsection{Architecture of the Stratified Hypergraph Ontology}
\label{subsec:hypergraph_architecture}
Unlike flat knowledge graphs that force complex relationships into binary triples and thereby fragment context, \ours adopts a Stratified Hypergraph Ontology to achieve a unified representation of enterprise data and business logic. This stratified design decouples stable data entities from evolving semantics, supporting iterative ontology refinement without requiring expensive pipeline engineering or physical schema changes.

\paragraph{Graph Layer (Virtual Entity-Relational Substrate).}
The Graph Layer consists of Nodes $\mathcal{V}$ and Edges $\mathcal{E}$ representing binary relational structures. Each node $v \in \mathcal{V}$ maps to a curated \textit{entity proxy} (e.g., a semantic view of a database table or API endpoint) rather than storing raw data. This design implements the \textit{data virtualization} paradigm from Virtual Knowledge Graphs (VKG) \cite{Xiao2019Virtual}, where data remains in-place and is accessed through declarative mappings. Core node attributes include: \texttt{id}, \texttt{name}, \texttt{description}, and \texttt{schema} (enriched with field-level annotations to guide the LLM reasoning engine). Edges $\mathcal{E}$ represent deterministic binary associations, such as cross-system foreign-key relationships, to support multi-hop data acquisition.

\paragraph{Hyperedge Layer ($n$-ary Semantic Ontology).}

A hyperedge $e$ acts as a higher-order container binding graph node subsets $\mathcal{V}_e \subseteq \mathcal{V}$ to localized \textbf{soft (natural language) axioms} $\mathcal{R}_e \subseteq \mathcal{R}$. Implemented within the \texttt{semantic\_details} field, these axioms express business conditions, anomaly triggers, and tool-use boundaries in a unified format. Functionally, \textbf{Declarative Hyperedges} ground admissible business states and cross-system relationships within the factual topology of the Graph Layer. Drawing inspiration from modular agentic skills \cite{wang2023voyager,schick2024toolformer}, \textbf{Procedural Hyperedges} build upon these declarative anchors to encapsulate reusable reasoning workflows and execution protocols. Ultimately, this human- and machine-interpretable representation provides the explicit structural scaffolding required for the orchestration LLM to navigate complex branching logic and multi-step validation with high fidelity.

Beyond these connecting nodes and soft axioms, a hyperedge $e$ is structurally defined by its \texttt{title} and \texttt{description}, which will be used later for retrieval. In addition, the \texttt{related\_hyperedges} and \texttt{attachments} define ontological dependencies and paths to executable artifacts (e.g., Python scripts). Hyperedges are also partitioned by governance into \textbf{Global} and \textbf{Scoped} types, ensuring architectural extensibility where role-specific protocols can be introduced without altering the underlying shared graph substrate.

\paragraph{Unified Heterogeneous Graph Structure.}
\label{subsec:hyperedge_association}
Following the principle of bipartite incidence representations \cite{bretto2013hypergraph}, we model hyperedges as first-class nodes within the architecture. This transformation consolidates the stratified ontology into a unified heterogeneous graph, formally parameterized by a tuple of five elements: \textbf{Base Graph Nodes} $\mathcal{V}$ and \textbf{Binary Edges} $\mathcal{E}$ representing entity proxies and their structural links; \textbf{Hyperedge Nodes} $\mathcal{V}_{\mathcal H}$ that model the hyperedges themselves; \textbf{Incidence Edges} $\mathcal{E}_{\mathcal V, \mathcal H}$ that define the membership connections between base graph nodes and hyperedge nodes; and \textbf{Inter-Hyperedge Edges} $\mathcal{E}_{\mathcal H}$ connecting hyperedge nodes via \texttt{related\_hyperedges} fields into a cognitive navigational network. This provides a standardized search space for both high-level semantic exploration and precise data retrieval.

\subsection{Ontology Tools for Access, Management, and Execution}
\label{sec:hypergraph_tools}

To bridge the LLM Reasoning Engine with physical data execution, \ours provides several core tool tiers---Graph Data Access, Topology Exploration, Hyperedge Retrieval, and Hyperedge Building---as well as auxiliary execution primitives. Together, these tools support transitions between localized evidence access, topology-guided generalization, and ontological-unit management, while ensuring that intermediate reasoning traces are provenance-aware and auditable.

\paragraph{Graph Data Access Tools.} 
This tool extracts physical evidence from the data substrate\footnote{Results are returned directly if small, or persisted as referenced artifacts if large.}. To strictly enforce enterprise governance, all operations are wrapped in a permission hook that validates user privileges and applies data-level access controls (e.g., truncating unauthorized fields or rejecting requests). Authorized access utilizes two modes:

\noindent\textbf{(1) Direct Access}: Executes LLM-generated structured queries (e.g., SQL, RESTful APIs) for deterministic retrieval.

\noindent\textbf{(2) Topology-Driven Access}: Operationalizes multi-hop gathering across data nodes. Given an input graph node subset and constraints (e.g., \texttt{SalesOrder.id=123}), it first verifies graph-layer connectivity to preclude hallucinated retrieval paths. Then the tool implements federated retrieval by sequentially propagating join keys along the validated subgraph, orchestrating localized queries to seamlessly bridge heterogeneous systems.

\paragraph{Hyperedge Retrieval Tools.} \label{subsec:hyperedge_retrieval}
To balance low-latency execution with flexible semantic search, \ours manages hyperedges via two mechanisms that enforce lazy context expansion:

\noindent\textbf{(1) Passive Activation}: For queries $q$ containing exact hyperedge titles or aliases, \ours uses multi-pattern string matching (e.g., Trie or Aho--Corasick \cite{fredkin1960trie,aho1975efficient}) for zero-latency detection. It injects a lightweight hyperedge summary (\texttt{title} and \texttt{description}) into the initial context $C$, deferring full \texttt{semantic\_details} to lazy loading.

\noindent\textbf{(2) Active Retrieval}: For complex queries lacking direct matches, the LLM generates searching queries $q_{search} = \text{LLM}_{plan}(q)$ based on the task goal. A hybrid retrieval strategy utilizes $q_{search}$ and hyperedge summaries, combining dense embeddings (e.g., BGE \cite{2025bge}) and sparse BM25 signals to rank candidates. Hyperedge summaries whose similarities exceed a composite threshold $\tau$ are returned to the LLM. 

The LLM then decides whether to invoke a detail-read operation for full \texttt{semantic\_details} and \texttt{attachments}. Crucially, this retrieval is not a single-pass operation; the reasoning engine can invoke the tool iteratively, dynamically pulling additional hyperedges across multiple reasoning steps as new intermediate findings or requirements emerge.

\paragraph{Topology Exploration Tools.} 
To navigate the stratified ontology beyond initially retrieved hyperedges while preventing context bloat, this tool offers two targeted traversal modes:
\textbf{(1) Local Adjacency Inspection}: Restricted to base graph nodes, this mode retrieves semantic descriptions and immediate binary relationships for specified entities. This enables safe, single-hop context expansion without uncontrolled recursive materialization.
\textbf{(2) Bounded Path Discovery}: Operating across all node types (both base graph nodes and hyperedges), this mode executes constrained traversal (e.g., breadth-first search) to autonomously discover viable multi-hop structural correlations between disjoint entities.
By orchestrating both modes, the LLM dynamically traces cross-system logic and valid graph data access queries---for example, peeking at a \texttt{Sales Order}'s binary neighbors before invoking path discovery through intermediate nodes (e.g., \texttt{Delivery Request}) to locate a final \texttt{Order Fulfillment} record.

\paragraph{Hyperedge Building Tool.} 
To support secure, continuous evolution without model retraining or expensive ontology building pipelines, \ours employs an adaptive construction framework enabling human-AI collaboration. While some declarative hyperedges are auto-instantiated directly from graph nodes during initial system setup, this tool empowers domain experts to dynamically create and refine both hyperedge types on demand. Users can iteratively adjust declarative constraints in \texttt{semantic\_details}. For procedural hyperedges, the tool facilitates LLM-assisted co-creation: the LLM drafts execution scripts from natural language intents, which experts then validate and finalize. This human-in-the-loop approach ensures scalable, verifiable ontology expansion.

This establishes a \textbf{self-improving cognitive loop}. While novel queries rely on zero-shot topology exploration, successful multi-hop reasoning traces for high-frequency tasks can be distilled into reusable procedural hyperedges. This progressively shifts the system from flexible exploration to highly efficient, deterministic protocol reuse. Finally, to ensure structural integrity, construction strictly enforces \textbf{role-based access control}: global hyperedges require root-level review, whereas tenant-scoped hyperedges isolate local logic, requiring explicit approval for any cross-scope dependencies.

\paragraph{Terminal and Code Execution Tools.}
These tools provide execution primitives for the operational steps referenced within \texttt{semantic\_details} or encoded in \texttt{attachments} for procedural hyperedges. They serve as the critical bridge that translates the soft protocols of a procedural hyperedge into actionable system operations, enabling the dynamic integration of external workflows and real-time automation. To ensure enterprise-grade security and prevent malicious system-level intrusions \cite{lu-etal-2025-chain}, all script executions and file system interactions are strictly confined within isolated sandbox environments.

\subsection{Agentic Reasoning Loop on the Stratified Hypergraph Ontology}\label{subsec:ai_engine}

Having established the Stratified Hypergraph Ontology, we now detail how \ours employs an LLM as its core engine to orchestrate a dynamic, evidence-driven agentic reasoning loop. Given an input query $q$, the system initializes an evolving context state $\mathcal{C}$. By structurally grounding the LLM's reasoning process within the ontology, each tool invocation explicitly enriches $\mathcal{C}$ with verifiable factual anchors. This mechanism guarantees adaptive planning and end-to-end auditability as the reasoning engine iteratively executes the following phases:
\begin{itemize}
    \item \textbf{Context Initialization:} A lightweight matching layer scans $q$ to detect relevant hyperedges (detailed in Section~\ref{subsec:hyperedge_retrieval}) or domain terms. These compact semantic hints are injected into the initial LLM context $\mathcal{C}$, deliberately avoiding the eager loading of full hyperedge details to conserve token space.

    \item \textbf{LLM Reasoning and Tool Selection:} Conditioned on the current context state $\mathcal{C}$ and historical tool outputs, the LLM decomposes the current objective into immediate actionable sub-tasks, formulating a local execution plan for the current iteration. Conceptually, the agent dynamically selects from the specialized tool tiers (defined in Section~\ref{sec:hypergraph_tools}) to: (i) invoke the \textbf{Hyperedge Retrieval Tool} to actively search for relevant hyperedge summaries, and subsequently make autonomous decisions to lazily load full \texttt{semantic\_details} for promising candidates; (ii) utilize the \textbf{Graph Data Access Tool} to extract physical evidence through direct queries or topology-driven federated retrieval; (iii) employ the \textbf{Topology Exploration Tool} to inspect graph nodes' adjacency or discover unmapped reasoning pathways when retrieved hyperedges are insufficient; or (iv) execute auxiliary primitives (e.g., safe terminal operations) through the \textbf{Terminal and Code Execution Tools}. These invocations can be generated simultaneously or fluidly interleaved.
    
    \item \textbf{Evidence Acquisition and Context Revision:} Upon receiving the execution results, \ours appends the newly acquired verifiable evidence directly to the context state $\mathcal{C}$\footnote{When the length of context $\mathcal{C}$ exceeds a predefined threshold, a system-level context compaction step is triggered to retain essential factual traces.}. This linearly expanded state then feeds directly back into the reasoning phase, ensuring that the agent's subsequent tool selections strictly adhere to the latest findings and the business constraints encoded within the active hyperedges.

    \item \textbf{Auditable Termination:} The iterative loop terminates when the LLM determines that the accumulated evidence in $\mathcal{C}$ is sufficient to definitively resolve the input query $q$. The final response $r$ is directly synthesized from, and firmly grounded in, this transparent and auditable evidence trace.
\end{itemize}

\paragraph{Typical Structured Multi-hop Reasoning Paths.}
While the core reasoning loop is unified, \ours realizes structured multi-hop reasoning through two primary execution paths, supported by intra- and inter-hyperedge navigation. \emph{Path A (Procedural Hyperedge-Guided)} relies on intra-hyperedge composition: the LLM actively retrieves and lazily loads a relevant procedural hyperedge whose \texttt{semantic\_details} encapsulate a reusable multi-hop protocol over graph nodes and the declarative anchors attached to them. For example, a procedural hyperedge for order fulfillment blockage root-cause analysis can prescribe an ordered evidence-acquisition program spanning \texttt{Contract Signing} (BPM), \texttt{Delivery Request} (ERP), \texttt{Outbound Delivery} (WMS), and supplier-side \texttt{Purchase Order} traces (SRM), strictly adhering to the conditional branches defined by the active soft axioms. 
Conversely, \emph{Path B (Topological Ontology Exploration)} serves as a fallback for long-tail queries lacking procedural hyperedges. Anchoring to declarative hyperedges, the LLM orchestrates the dual-mode Topology Exploration Tool---dynamically alternating between local adjacency inspection and bounded path discovery---to autonomously navigate the underlying hypergraph and resolve novel reasoning pathways.
Alongside utilizing the \textbf{Topology Exploration Tool} for navigation, the LLM could also leverage the \texttt{related\_hyperedges} field to query semantic neighbors, recursively load relevant successors, and assemble an interpretable high-level chain (e.g., $e_1 \rightarrow e_2 \rightarrow \dots \rightarrow e_k$) tailored to the dynamically evolving query context.

\section{Experiments}\label{sec:experiments}
\subsection{Experimental Setup}

\subsubsection{Datasets}
We extracted real-world business data from the Order Fulfillment (OF) process of \supcon. To prevent irreproducible results caused by continuous production updates, we synchronized static snapshots of 20 tables to local databases. Spanning heterogeneous systems (BPM, ERP, SRM, and WMS), this provides a high-fidelity foundation for evaluating multi-hop analysis tasks. We construct two datasets sharing the same database snapshots but differing in queries and ground-truth generation:
(1) \emph{Order Fulfillment Blockage Root-Cause Analysis (\dofba):} 
95 standardized questions following the template ``Why has contract [X] not completed fulfillment?''. Answering requires chaining evidence across all systems to produce a grounded diagnosis. As this reflects an actual operational bottleneck, the ground truth was rigorously established by domain experts who manually investigated each contract across the heterogeneous systems.
(2) \emph{Order Fulfillment Generalization (\dofgeneral):}
160 general and diverse questions categorized by required reasoning span. Span is operationalized by the number of distinct data nodes involved, measured as $k = |\texttt{nodes}| \in \{0,1,2,3,>3\}$. To construct the ground truth for these diverse scenarios, domain experts engaged in multi-turn interactions with an AI assistant, actively verifying retrieved evidence, correcting reasoning trajectories, and refining the final conclusions to align with accepted enterprise statistical standards and operational definitions.

\subsubsection{Evaluation Metrics \& Configuration}
We evaluate three core metrics: \textbf{Accuracy}, \textbf{Tool turns} (iterations executed), and total \textbf{Tokens} consumed.
To validate accuracy, we adopt the LLM-as-a-judge framework \cite{gu2024survey} followed by human expert verification, evaluating the final conclusion alongside its supporting tool evidence. For \dofgeneral{}, we apply a three-way label (\emph{correct}/\emph{partially correct}/\emph{incorrect}) to accommodate definitional nuances, reporting \emph{correct} and \emph{correct}+\emph{partially correct} accuracies. For \dofba{}, accuracy is strictly binary (\emph{correct} vs.\ \emph{incorrect}).

\ours runs with a maximum budget of 50 tool turns. Unless stated otherwise, we use GPT-5 as the default LLM backbone, but also evaluate DeepSeek-3.2, Qwen-3.5-27B, and Qwen-Next 80B to quantify model-capability effects.
Declarative hyperedges are auto-extracted and refined to bind soft business axioms to established graph nodes. Crucially, we authored a dedicated procedural hyperedge for \dofba{} to construct the validated fulfillment-diagnosis protocol, explicitly aligning it with Path A (Procedural Hyperedge-Guided) execution. Lacking such predefined protocols, \dofgeneral{} naturally defaults to Path B (Topological Ontology Exploration) for autonomous multi-hop discovery.

\subsubsection{Ablation and Baselines}

To isolate our hypergraph representation's advantages, we compare \ours against ablations and baselines. For fair evaluation, all baselines receive the identical business axioms $\mathcal{R}$ (Section \ref{sec:problem_formulation}) as prompts in their initial context.
\begin{itemize}
    \item \textbf{HEAR (Complete)}: The proposed method with both declarative anchors and procedural protocols available.
    \item \textbf{HEAR (Declarative)}: Only declarative hyperedges are available; procedural hyperedges are removed.
    \item \textbf{Table-RAG}: Hyperedges removed. Uses a highly scalable, multi-granularity index over schema metadata via LlamaIndex \cite{llamaindex_sql_docs} to generate and execute SQL.
    \item \textbf{Table list}: Hyperedges removed. Relies autonomously on a flat, semantically processed list of table identifiers and interface descriptions in the initial context.
    \item \textbf{CSR-RAG}: A state-of-the-art multi-table retrieval architecture \cite{singh2026csrrag} that constructs a schema hypergraph to generate SQL, entirely outside the \ours agentic loop.
\end{itemize}

\subsection{Evaluation on \dofba Dataset}
Evaluating \ours on \dofba{} (Table~\ref{tab:ofba-of100-ablation}), HEAR (Complete) achieves a dominant 94.7\% accuracy. Ablation reveals procedural hyperedges critically constrain exploration: removing them drops accuracy by 20.5 points (to 74.2\% for HEAR Declarative) and drastically inflates search costs (from 6.3 turns/38.1K tokens to 14.9 turns/179.3K tokens). Yet, this declarative-only variant still significantly outperforms Table-RAG (52.6\%), Table list (51.5\%), and CSR-RAG (11.3\%), proving axiomatic structure alone establishes a superior foundation. Error analysis reinforces this: lacking global state tracking, Table-RAG conflates disconnected evidence, while CSR-RAG's collapse confirms static pipelines outside an iterative loop inherently fail at complex multi-hop reasoning.

\begin{figure}[t]
    \centering
    \includegraphics[width=\linewidth]{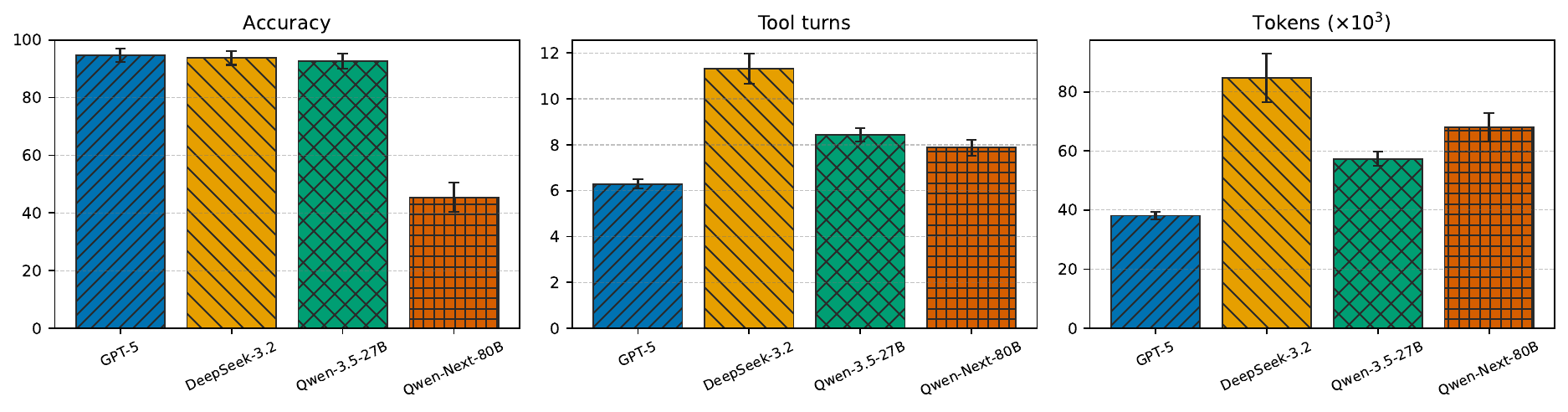}
    \caption{Accuracy and efficiency on \dofba{} across LLM backbones when both declarative and procedural hyperedges are available.}
    \label{fig:ofba-of100}
\end{figure}

Comparing LLM backbones under complete HEAR (Figure~\ref{fig:ofba-of100}), the proprietary GPT-5 (94.7\%) and DeepSeek-3.2 (93.8\%) narrowly lead the open-weight Qwen-3.5-27B (92.7\%), while Qwen-Next-80B collapses (45.4\%). Although GPT-5 is the most efficient (6.29 turns, 38.1K tokens), deploying Qwen-3.5-27B demonstrates that an open-weight model can achieve proprietary-level accuracy with exceptional cost-efficiency. In contrast, DeepSeek-3.2 incurs materially higher costs (84.7K tokens) due to a conservative, verification-heavy style. Finally, residual errors across the top models cluster around undocumented enterprise anomalies (e.g., supplier dropship logic), indicating that long-tail workflow ambiguity—not generic reasoning capacity—is the primary bottleneck.

\begin{table}[t]
    \centering
    \caption{Comparison on \dofba{} to evaluate the importance of hyperedges. Here HEAR (Complete) exposes both declarative and procedural hyperedges, whereas HEAR (Declarative) keeps only declarative hyperedges. Accuracy is binary (correct vs. incorrect).}
    \label{tab:ofba-of100-ablation}
    \begin{tabular}{lllccc}
\toprule
Category & Method & Hyperedges & Accuracy (\%) & Tool turns & Tokens ($\times 10^3$) \\
\midrule
\multirow{2}{*}{} & \multirow{2}{*}{HEAR} & Complete & 94.7 & 6.3 $\pm$ 0.2 & 38.1 $\pm$ 1.3 \\
 &  & Declarative & 74.2 & 14.9 $\pm$ 0.4 & 179.3 $\pm$ 7.1 \\
\midrule
\multirow{2}{*}{\textbf{Internal Baseline}} & Table-RAG & No & 52.6 & 10.8 $\pm$ 0.5 & 78.5 $\pm$ 5.2 \\
 & Table list & No & 51.5 & 4.9 $\pm$ 0.3 & 63.1 $\pm$ 5.0 \\
\midrule
\textbf{External Baseline} & CSR-RAG & — & 11.3 & 3.0 $\pm$ 0.0 & 16.7 $\pm$ 0.7 \\
\bottomrule
\end{tabular}

\end{table}

\subsection{Evaluation on \dofgeneral Dataset}

\begin{table}[t]
    \centering
    \caption{Comparison on \dofgeneral{} to evaluate the importance of hyperedges. Under Accuracy (\%), \emph{Correct} reports fully correct answers only, and \emph{+Partial} reports fully correct plus partially correct answers.}
    \label{fig:ofgeneral-eval3-ablation}
    \begin{tabular}{lllcccc}
\toprule
\multirow{2}{*}{Category} & \multirow{2}{*}{Method} & \multirow{2}{*}{Hyperedges} & \multicolumn{2}{c}{Accuracy (\%)} & \multirow{2}{*}{Tool turns} & \multirow{2}{*}{Tokens ($\times 10^3$)} \\
& & & Correct & +Partial & & \\
\midrule
\multirow{2}{*}{} & \multirow{2}{*}{HEAR} & Complete & 37.1 & 88.7 & 7.3 $\pm$ 0.4 & 81.6 $\pm$ 8.5 \\
 &  & Declarative & 35.8 & 86.2 & 7.4 $\pm$ 0.3 & 76.7 $\pm$ 5.8 \\
\midrule
\multirow{2}{*}{\textbf{Internal Baseline}} & Table-RAG & No & 29.6 & 85.5 & 6.7 $\pm$ 0.3 & 40.8 $\pm$ 3.7 \\
 & Table list & No & 28.9 & 88.7 & 4.5 $\pm$ 0.2 & 35.1 $\pm$ 3.0 \\
\midrule
\textbf{External Baseline} & CSR-RAG & — & 6.3 & 72.2 & — & 8.1 $\pm$ 0.4 \\
\bottomrule
\end{tabular}

\end{table}

\begin{figure}[t]
    \centering
    \includegraphics[width=\linewidth]{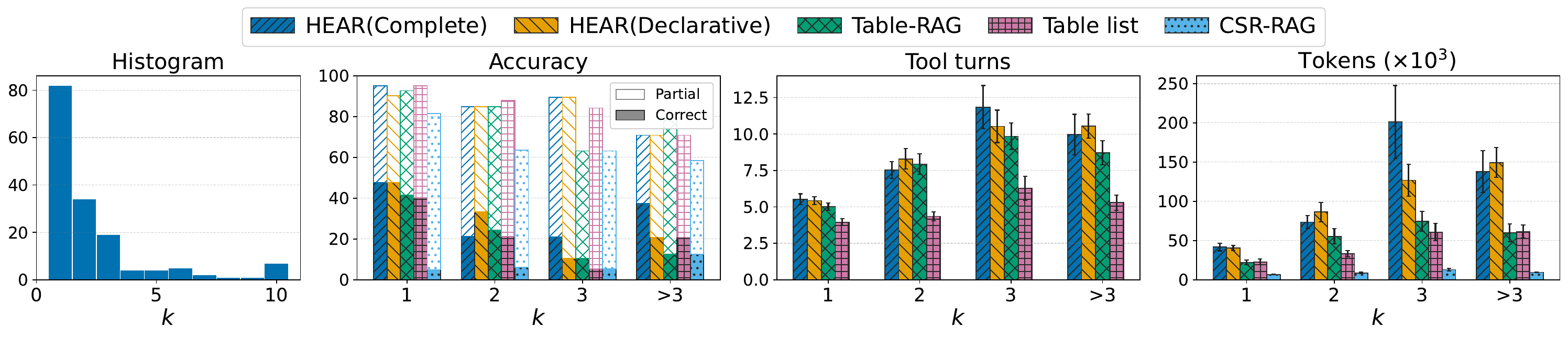}
    \caption{Comparison on \dofgeneral{} by node-count bucket (defined by $k = |\texttt{nodes}|$ as a proxy for required reasoning span). The stacked bars report \emph{Correct} and \emph{+Partial} accuracies.}
    \label{fig:ofgeneral-ablation-by-complexity}
\end{figure}

\begin{figure}[t]
    \centering
    \includegraphics[width=\linewidth]{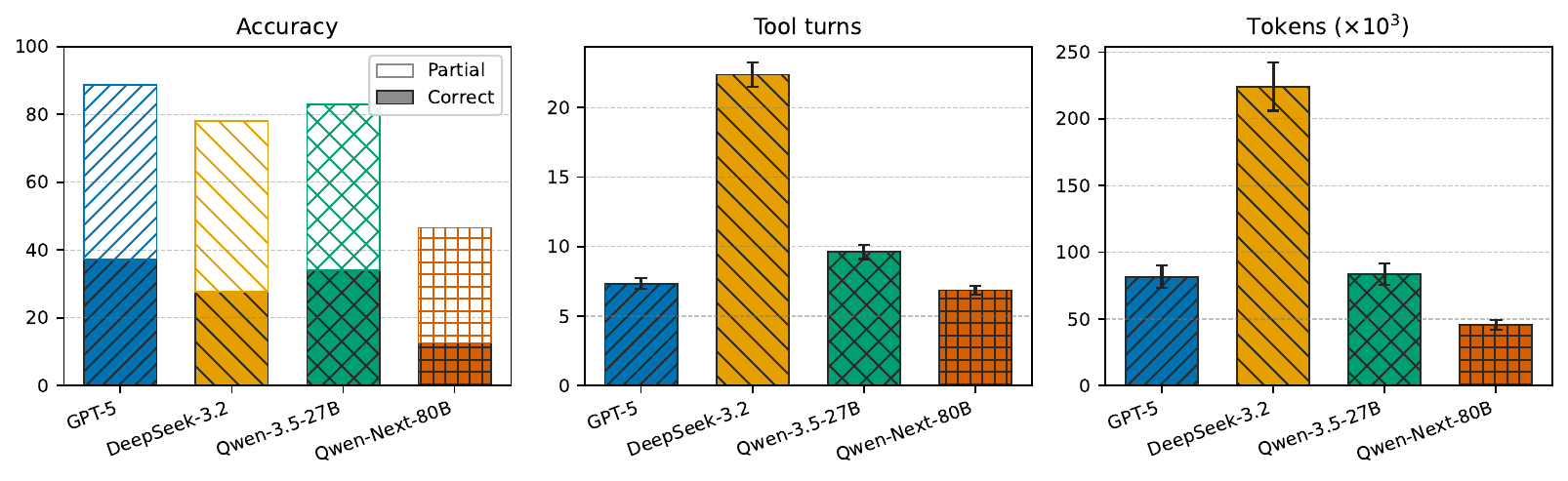}
    \caption{LLM backbone comparison on \dofgeneral{}.}
    \label{fig:ofgeneral-eval3}
\end{figure}

We next evaluate on \dofgeneral{} (Table~\ref{fig:ofgeneral-eval3-ablation}). Unlike \dofba{}, this diverse dataset lacks procedural hyperedges targeted at specific questions, forcing Path B (Topological Ontology Exploration) for multi-hop reasoning. While baselines appear competitive under the lenient \emph{Correct+Partial} metric (both HEAR Complete and Table list hit 88.7\%), the strict \emph{Correct} metric shows clear separation. HEAR (Complete) peaks at 37.1\%. Notably, HEAR (Declarative) also performs strongly (35.8\% fully correct), proving the value of declarative axioms even without predefined protocols. Both outperform Table-RAG (29.6\%), while CSR-RAG performs poorest (6.3\%) because its lack of an agentic reasoning loop severely limits multi-hop evidence gathering. \ours pays a higher token cost than Table-RAG (81.6K vs. 40.8K tokens) for this capability, a reasonable trade-off when prioritizing correctness.

By-complexity results (Figure~\ref{fig:ofgeneral-ablation-by-complexity}) further demonstrate this exploration dynamic under the \emph{Correct+Partial} metric. As reasoning spans grow and predefined routes vanish, Path B execution remains competitive across all buckets: HEAR ties or leads the top scores at $k{\le}2$ (95.1\% and 84.8\%), leads clearly at $k{=}3$ (89.5\%), and remains tied at $k{>}3$ (70.8\%). The clearest separation appears in the mid-complexity $k{=}3$ bucket, where HEAR strictly outperforms Table list (84.2\%) and Table-RAG (63.2\%). Constructing longer reasoning chains naturally raises execution costs: turns increase from 5.52 ($k{=}1$) to 11.84 ($k{=}3$), while tokens rise from 41.7K to 201.1K. Although baselines like Table-RAG use fewer tokens on harder buckets, their weaker accuracy reinforces our trade-off: \ours accepts moderate overhead for reliable multi-hop reasoning.

Comparing LLM backbones under complete HEAR (Figure~\ref{fig:ofgeneral-eval3}), GPT-5 (88.7\%) narrowly leads Qwen-3.5-27B (83.0\%). Echoing \dofba{} results, Qwen stands out as a highly competitive open-weight alternative that reaches near-frontier accuracy. DeepSeek-3.2 is weaker in both effectiveness (78.0\%) and efficiency (22.4 turns, 224.1K tokens vs. GPT-5's 7.3 turns, 81.6K tokens) due to a verification-heavy style. Overall, robust hypergraph grounding makes Qwen-3.5-27B a highly practical open-weight choice, while GPT-5 retains the strongest overall profile on complex questions.

\subsection{Maintenance Overhead and Practicality}

The Stratified Hypergraph Ontology minimizes knowledge maintenance overhead. Across 20 tables, declarative hyperedges are instantiated in seconds via automated metadata extraction, requiring only hours for experts to refine \texttt{semantic\_details}. Constructing procedural hyperedges demands targeted upfront investment; e.g., formalizing ``order fulfillment blockage analysis'' requires 1--2 days of human--AI co-creation.

However, this yields compounding gains. Conventionally, diagnosing 200 blockages takes ~20 person-days manually. Grounded by procedural hyperedges, \ours executes this workload in under 10 hours with verified correctness. By interfacing with RPA endpoints to trigger corrective actions, \ours translates minimal maintenance into a paradigm shift: from reactive diagnostics to proactive automation.
\section{Conclusion and Future Work}
\label{sec:conclusion}

In this work, we introduced \textbf{\ours}, a Hypergraph Enterprise Agentic Reasoner that grounds LLMs within a Stratified Hypergraph Ontology. 
By resolving heterogeneous data dependencies and modeling $n$-ary business constraints to enforce traceable multi-hop reasoning, \ours enables high-fidelity, audit-compliant execution for complex enterprise tasks.
Evaluations on real-world supply-chain data demonstrate that explicit hypergraph semantics secure strict accuracy and adaptive execution efficiency.

Future research will focus on scaling and hardening this cognitive infrastructure through: (1) automated hypergraph entropy control to prune redundancy and retrieval noise; (2) graph-enhanced embeddings for more robust semantic hyperedge retrieval; and (3) seamless alignment with established ontology standards (e.g., RDF, OWL) to leverage standardized knowledge schemas and enhance cross-system interoperability. Ultimately, we aim to provide an auditable, scalable, and self-improving foundation for next-generation enterprise intelligence.

\newpage

\section*{Acknowledgment}
We thank Yao Yao, Yongfeng Zhang, Teng Wang, Ao Sun and Wei Ge for their assistance with database setup, hyperedge semantic drafting, dataset construction and contributions to system implementation. We also express our gratitude to the Order Management Department for supplying domain-specific business constraints and verifying the system in real-world production. 

\section*{Declaration of Use of Generative AI}
Gemini, GPT, and Kimi were utilized to generate sections of this work, including underlying source code, bibliographic metadata, and drafted prose for the abstract, introduction, and experiments. The core intellectual contributions, system architecture, and experimental design were authored entirely by the human researchers. These models functioned strictly as assistants; the authors critically reviewed, edited, and verified all AI-generated content, assuming full responsibility for the accuracy, originality, and integrity of the entire work.


\section*{Supplemental Material Statement}
Due to strict proprietary policies and non-disclosure agreements (NDAs) with the industrial partner, neither the raw datasets (\dofba and \dofgeneral) nor the integrated source code can be publicly released. The datasets contain sensitive production data from core enterprise manufacturing systems (ERP, SRM, etc.), and the source code involves proprietary production interfaces. However, to ensure reproducibility and facilitate the verification of our results, this paper is designed to be fully self-contained. Extended details regarding the proposed methodology\todo{comprehensive pseudo-code,} and further experimental configurations and findings are thoroughly documented in the Annexes of this submission.

\bibliographystyle{plain}
\bibliography{combined_ref}

\appendix

\section{Extended Problem Formulation Details}
\label{app:data_anomalies}

\subsection{Semantic Interoperability Issues in Enterprise Data Integration}

In real-world production environments, achieving semantic interoperability is severely bottlenecked by a myriad of implicit data and process anomalies. We identify the following common issues that plague enterprise data integration:

\begin{itemize}
    \item \textbf{Semantic Ambiguity in Schemas}: Table and field names often lack clear semantics (e.g., tables using German abbreviations or Pinyin acronyms).
    \item \textbf{Field-Level Equivocation}: Similar field names obscure distinct business meanings (e.g., multiple ``effective dates'' coexisting without explicit definitions).
    \item \textbf{Opaque Statistical Metrics}: Calculation logic and aggregation rules are frequently undocumented, relying heavily on tribal knowledge among employees.
    \item \textbf{Cross-Table Identifier Inconsistency}: The same business entity may be labeled differently across tables (e.g., a contract identifier named \texttt{contract\_no} in one table and \texttt{contract\_code} in another).
    \item \textbf{Format and Encoding Discrepancies}: Cross-system identifiers often suffer from inconsistent formatting, such as material codes being zero-padded in one system but not in another.
\end{itemize}

\section{Extended Method Details}

\subsection{Graph Layer Node Attributes}
Each node $v \in \mathcal{V}$ in the Graph Layer possesses core attributes designed to secure channels and align semantics during LLM planning:
\begin{itemize}
    \item \texttt{name} \& \texttt{id}: Globally unique identifiers (e.g., ``table:wms\_inventory'').
    \item \texttt{description}: Natural language descriptions capturing the node's precise business meaning.
    \item \texttt{source}: The physical topological origin of the data source (e.g., WMS).
    \item \texttt{schema}: Structured schema definitions (e.g., simplified SQL DDL). To optimize the LLM's token context window, we utilize field cleaning and high-level semantic annotations rather than enumerative full-field outputs.
    \item \texttt{query\_template}: Parameterized query templates (e.g., SQL queries or REST API payloads) for securely retrieving instance data.
\end{itemize}

\subsection{Hyperedge Attributes}
Key attributes of a hyperedge $e \in \mathcal{H}$ include:
\begin{itemize}
    \item \texttt{title}: A unique identifier and natural language specification of the hyperedge.
    \item \texttt{description}: A summary of the hyperedge's task and core logic.
    \item \texttt{nodes $\mathcal{V}_e$}: The subset of graph nodes defining the required data scope.
    \item \texttt{semantic\_details}: The core cognitive carrier binding the business constraints to the graph nodes. It specifies business conditions, multi-step reasoning, domain validity checks, anomaly triggers, and canonical tool-use procedures.
    \item \texttt{related\_hyperedges}: References to other hyperedges expressing dependencies, reasoning fallbacks, or next-step actions.
    \item \texttt{attachments}: Optional executable scripts (e.g., Python) for automating analysis steps.
\end{itemize}

\section{Extended Experimental Details}

\subsection{Dataset Profiles and Reasoning Span Definitions}
\label{app:dataset_details}

For the \dofgeneral{} dataset, we operationalize reasoning span by the number of distinct data nodes involved in a question ($k = |\texttt{nodes}|$). The specific buckets and corresponding examples are defined as follows:

\begin{itemize}
    \item \textbf{$k{=}1$}: Single-node retrieval and reasoning. 
    \textit{Examples: (1) Top 10 customers by contract volume; (2) A breakdown of the most frequently used contract types.}
    \item \textbf{$k{=}2$} and \textbf{$k{=}3$}: Require retrieval and reasoning across 2--3 nodes. 
    \textit{Examples: (1) Which suppliers are associated with order O123? (2) Count of contracts with over-receipt (site receipt > delivery request) for direct-ship physical items.}
    \item \textbf{$k{>}3$}: Require retrieval, multi-hop reasoning, and calculation involving more than 3 nodes. 
    \textit{Examples: (1) Order O123 has generated a delivery request but has not yet left the warehouse; what are the possible causes? (2) What are the top 5 pending approvers with the highest number of blocked lines for Delivery Requests under review, including their individual blocked line counts and the overall total?}
\end{itemize}

\subsection{Two-Stage Verification Protocol and Label Spaces}
\label{app:metric_details}

To ensure rigorous validation of the \textbf{Accuracy} metric, we employ a two-stage verification protocol: an initial assessment using an LLM-as-judge approach~\cite{gu2024survey}, followed by a final verification conducted by human experts. The judged output includes both the final conclusion and the organized supporting evidence gathered from tool outputs, meaning correctness is evaluated on the grounded answer package rather than on a free-form claim alone. 

For \dofgeneral{}, we use a three-way final-correctness label space (\emph{correct} / \emph{partially correct} / \emph{incorrect}) to account for acceptable definition-level differences (e.g., alternative aggregation windows or statistical standards). For \dofba{}, the intermediate ``partially correct'' outcome is negligible in our runs as a contract will basically have only one root cause of unfulfillment, so accuracy is effectively binary (\emph{correct} vs.\ \emph{incorrect} conclusions).

\subsection{Internal Baseline Implementation Context}
\label{app:baseline_details}

To rigorously ablate the contribution of the Stratified Hypergraph Ontology, we designed internal baselines that share the identical agentic loop, tool set, and turn budget as \ours, but are stripped of hyperedge semantics:

\begin{itemize}
    \item \textbf{Table-RAG}: To provide the LLM with relevant table context without triggering the index explosion typical of raw content indexing (e.g., T-RAG \cite{zou2025trag}), we implement a strong schema-aware Table-RAG. Fusing these concepts with LlamaIndex \cite{llamaindex_sql_docs}, we build an index over schema metadata (tables, columns, types, and short descriptions) across heterogeneous databases. Retrieval yields lightweight schema pointers via a mapping function $f$ rather than raw table content, forcing the LLM to generate and execute SQL absent structural guidance.
    \item \textbf{Table list}: The initial reasoning context provides a flat list of table identifiers and short interface descriptions. This seemingly naive injection of semantically processed names yields a highly condensed information advantage, forming a remarkably strong baseline against which the value of explicit hypergraph topology can be measured.
\end{itemize}

\subsection{\dofba{} Error Modes and Comparative Analysis}
\label{app:of100-error-modes}

The evaluation on the \dofba{} dataset reveals distinct error patterns across different retrieval settings and LLM backbones, underscoring the challenges of multi-hop enterprise reasoning.

\paragraph{Baseline and Ablation Analysis.} 
The critical role of the complete hypergraph structure is validated by comparing \ours{} against various baselines and ablations. Without the topological guidance provided by hyperedges, the error rates increase dramatically. For the GPT-5 backbone, the number of errors rises from 5 in the HEAR (Complete) setting to 46 for Table-RAG and 86 for CSR-RAG. Similar performance degradation is observed for Qwen-3.5-27B, where errors increase from 7 to 52 and 86 in the respective baselines.

The distribution of error modes shifts significantly in these baselines. Specifically, failures related to missing Site Goods Receipts and general Order Fulfillment state confusion experience a massive surge, frequently exceeding 40--70 cases per baseline. This suggests that without hyperedge-based context, models struggle to maintain the global logic of the fulfillment chain, leading to hallucinations regarding the current state of a business object when faced with disconnected tabular evidence. For example, in one Table-RAG case, the ground-truth blockage was an upstream delivery application that had been submitted but not yet made effective, so the contract had not entered the outbound-delivery stage; the baseline instead concluded that outbound delivery had already completed and that the remaining issue was a missing site receipt. Furthermore, tracking Direct Shipment pathways becomes highly error-prone without structural guidance, jumping to over 30 errors. Ablating the hypergraph to only the declarative variant, HEAR (Declarative), also results in 30--50 errors. This proves that declarative relational links alone are insufficient for tracking long-span business processes that strictly mandate cross-system verification among BPM, ERP, SRM, and WMS.

\paragraph{Main \dofba{} Errors by LLM Backbone.} 
Under the full \ours{} configuration, denoted as HEAR (Complete), top-tier models demonstrate exceptional precision. Specifically, GPT-5, DeepSeek-3.2, and Qwen-3.5-27B only encounter 5, 6, and 7 failure cases, respectively. In contrast, the performance of the Qwen-Next-80B model drops significantly with 53 recorded errors under the same setting. 

The residual failures for high-performing models are not random but tightly clustered around a small subset of high-complexity business instances. Among the incorrect cases, the most frequent error modes involve missing Delivery Requests, complex Direct Shipment chains involving Vendor ASNs, and nuanced workflow state discrepancies. Qualitatively, these failures typically arise from a misinterpretation of specific blockage stages—for example, mistaking a missing upstream Delivery Request in the BPM system for a downstream synchronization failure between WMS (Outbound Delivery) and ERP (Financial Outbound). The high degree of overlap among different backbones on these residual cases (with several instances being marked incorrect by over 10 different systems) indicates that the difficulty lies in the intrinsic complexity of specific enterprise fulfillment traces across heterogeneous systems, rather than model-specific weaknesses.

\subsection{Additional Qwen-3.5-27B Baseline-Ablation Results}
\label{app:qwen-ablation-results}

The main text reports the GPT-5 baseline-ablation results. For completeness, we place the corresponding Qwen-3.5-27B versions here. These results verify that the advantages of hyperedge-structured reasoning are not tied to a proprietary backbone: even under a smaller open-weight model, HEAR (Complete) consistently produces the strongest accuracy, while ablations and retrieval-only baselines either lose correctness or fall back to less disciplined exploration.

\begin{table}[t]
    \centering
    \caption{Qwen-3.5-27B baseline ablation on \dofba{}.}
    \label{tab:appendix-ofba-qwen-ablation}
    \begin{tabular}{lllccc}
\toprule
Category & Method & Hyperedges & Accuracy (\%) & Tool turns & Tokens ($\times 10^3$) \\
\midrule
\multirow{2}{*}{} & \multirow{2}{*}{HEAR} & Complete & 92.7 & 8.4 $\pm$ 0.3 & 57.3 $\pm$ 2.4 \\
 &  & Declarative & 73.2 & 11.8 $\pm$ 0.5 & 105.9 $\pm$ 9.4 \\
\midrule
\multirow{2}{*}{\textbf{Internal Baseline}} & Table-RAG & No & 46.4 & 13.6 $\pm$ 0.6 & 122.6 $\pm$ 10.4 \\
 & Table list & No & 54.2 & 8.8 $\pm$ 0.4 & 48.9 $\pm$ 4.1 \\
\midrule
\textbf{External Baseline} & CSR-RAG & — & 9.5 & 3.0 $\pm$ 0.0 & 13.6 $\pm$ 0.8 \\
\bottomrule
\end{tabular}

\end{table}

On \dofba{} (Table~\ref{tab:appendix-ofba-qwen-ablation}), HEAR (Complete) reaches 92.7\% accuracy. Notably, HEAR (Declarative) establishes a robust foundation by achieving 73.2\% accuracy, which clearly exceeds Table-RAG (46.4\%), Table list (54.2\%), and CSR-RAG (9.5\%). Building upon this declarative base, adding procedural hyperedges yields a further 19.5-point accuracy jump while delivering a distinct efficiency gain: average tool turns drop from 11.8 to 8.4, and token usage decreases from 105.9K to 57.3K. Although Table list is slightly cheaper than HEAR (Complete) in tokens (48.9K vs.\ 57.3K), its accuracy remains far lower, confirming that concise schema exposure cannot replace procedural guidance for blockage diagnosis.

\begin{table}[t]
    \centering
    \caption{Qwen-3.5-27B baseline ablation on \dofgeneral{}.}
    \label{tab:appendix-ofgeneral-qwen-ablation}
    \begin{tabular}{lllcccc}
\toprule
\multirow{2}{*}{Category} & \multirow{2}{*}{Method} & \multirow{2}{*}{Hyperedges} & \multicolumn{2}{c}{Accuracy (\%)} & \multirow{2}{*}{Tool turns} & \multirow{2}{*}{Tokens ($\times 10^3$)} \\
& & & Correct & +Partial & & \\
\midrule
\multirow{2}{*}{} & \multirow{2}{*}{HEAR} & Complete & 34.0 & 83.0 & 9.6 $\pm$ 0.5 & 83.5 $\pm$ 8.2 \\
 &  & Declarative & 28.9 & 79.2 & 9.2 $\pm$ 0.5 & 79.3 $\pm$ 8.8 \\
\midrule
\multirow{2}{*}{\textbf{Internal Baseline}} & Table-RAG & No & 25.2 & 79.2 & 12.7 $\pm$ 0.9 & 154.3 $\pm$ 20.2 \\
 & Table list & No & 22.6 & 74.2 & 8.9 $\pm$ 0.5 & 58.8 $\pm$ 7.2 \\
\midrule
\textbf{External Baseline} & CSR-RAG & — & 3.2 & 57.0 & — & 9.7 $\pm$ 0.3 \\
\bottomrule
\end{tabular}

\end{table}

On \dofgeneral{} (Table~\ref{tab:appendix-ofgeneral-qwen-ablation}), the same trend appears under the stricter fully \emph{Correct} metric. HEAR (Complete) obtains 34.0\% fully correct accuracy and 83.0\% Correct+Partial accuracy, outperforming HEAR (Declarative) (28.9\% / 79.2\%), Table-RAG (25.2\% / 79.2\%), Table list (22.6\% / 74.2\%), and CSR-RAG (3.2\% / 57.0\%). The cost profile again reflects the nature of Path B exploration, but here tool turns and tokens should be read together: HEAR (Complete) does not uniformly minimize tool calls relative to every ablated setting, but it accepts a slightly higher overall computation—especially in token usage relative to external baselines (83.5K vs.\ 79.3K for HEAR (Declarative) and 58.8K for Table list)—because it continues assembling and validating multi-node evidence instead of stopping at shallow schema matches. In other words, the additional cost on \dofgeneral{} is not wasted exploration, but a task-appropriate investment in better-supported answer construction, which is precisely what yields the best strict accuracy.

\begin{figure}[t]
    \centering
    \includegraphics[width=\linewidth]{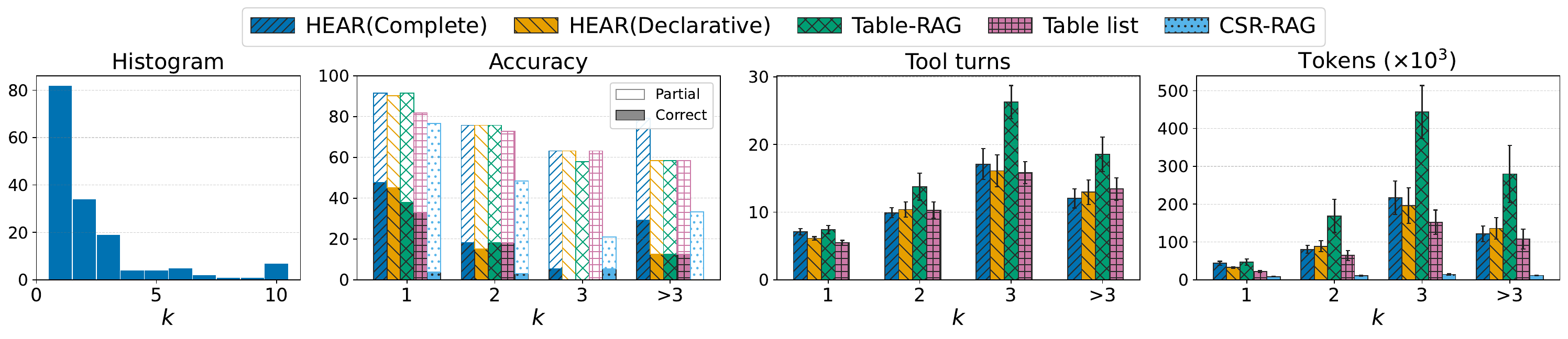}
    \caption{Qwen-3.5-27B baseline ablation on \dofgeneral{} by node-count bucket .}
    \label{fig:appendix-ofgeneral-qwen-ablation-by-complexity}
\end{figure}

The by-complexity breakdown (Figure~\ref{fig:appendix-ofgeneral-qwen-ablation-by-complexity}) further clarifies where the complete hyperedge structure matters. For simple $k{=}1$ questions, HEAR (Complete) and Table-RAG both reach 91.5\%, with HEAR (Declarative) close behind at 90.2\%. As the reasoning span grows, however, the benefit of complete hyperedges becomes more pronounced: at $k{=}3$, HEAR (Complete) reaches 63.2\%, compared with 57.9\% for Table-RAG and 21.1\% for CSR-RAG; at $k{>}3$, HEAR (Complete) rises to 79.2\%, while HEAR (Declarative), Table-RAG, and Table list all remain at 58.3\%, and CSR-RAG falls to 33.3\%. The cost curves show the same scaling effect. Both tool turns and token usage rise with reasoning span, with HEAR (Complete) increasing from 7.11 turns at $k{=}1$ to 17.11 at $k{=}3$, while tokens grow from 43.4K to 216.9K before settling at 121.5K for $k{>}3$. Thus, Qwen-3.5-27B reproduces the central conclusion of the main experiments: hyperedge-guided exploration is most valuable precisely when the question requires long-span, multi-system evidence assembly.

Taken together, the Qwen-3.5-27B results reproduce the same high-level pattern as GPT-5, but with a clearer contrast between two regimes. On \dofba{}, procedural hyperedges improve both accuracy and efficiency, reducing both tool turns and token cost relative to the declarative ablation. On \dofgeneral{}, they improve strict correctness at higher computational cost, with the extra burden appearing primarily in tokens and, for harder questions, also in longer tool-use chains. This distinction strengthens the overall claim of the paper: the value of hyperedges is not just that they increase accuracy, but that they allocate computation in a task-appropriate way.

\subsection{Additional Backbone-Specific Findings}
\label{app:backbone-findings}

Beyond the headline accuracy--cost comparisons in the main text, our traces suggest two recurring backbone-specific behavioral patterns.

\paragraph{Why DeepSeek-3.2 uses many more tool turns.}
On both \dofba{} and \dofgeneral{}, DeepSeek-3.2 is noticeably more tool-intensive than the other strong backbones. In our traces, this is not mainly because it explores dramatically broader search spaces, but because it more often re-checks intermediate artifacts before committing to the next step. Typical behaviors include reopening generated files, re-reading SQL outputs, and issuing additional verification-oriented tool calls after a seemingly sufficient intermediate result has already been obtained. This verification-heavy style can sometimes slightly improve final correctness, but it also increases latency and token consumption substantially. In other words, DeepSeek-3.2 behaves more conservatively inside the \ours loop: it spends more turns validating previous steps rather than advancing with a shorter execution chain.

\paragraph{Why Qwen-Next-80B does not outperform Qwen-3.5-27B.}
Our results show that a larger nominal model does not automatically yield better tool-grounded performance. Although Qwen-Next-80B has a larger advertised scale, it underperforms Qwen-3.5-27B on both datasets. A plausible explanation is architectural: the Qwen ``A3B'' naming convention is used for sparse MoE models with only a small subset of parameters activated per token, whereas the open-weight Qwen2.5 language models in the 27B/32B class are dense models according to the official Qwen release materials. Under \ours, this means the comparison is not simply ``bigger vs. smaller,'' but sparse-MoE scale versus a smaller dense backbone. In our runs, Qwen-3.5-27B more often follows concise and coherent action chains, whereas Qwen-Next-80B more frequently loses precision during tool-grounded execution. This is consistent with the broader observation that sparse MoE scale does not necessarily translate into stronger performance in tightly constrained, multi-step tool-use settings.

\end{document}